%% file: conference_101719.tex
\documentclass[a4paper,conference]{IEEEtran}
\IEEEoverridecommandlockouts
\usepackage{cite}
\usepackage{amsmath,amssymb,amsfonts}
\usepackage{algorithmic}
\usepackage{graphicx}
\usepackage{textcomp}
\usepackage{xcolor}
\def\BibTeX{{\rm B\kern-.05em{\sc i\kern-.025em b}\kern-.08em
    T\kern-.1667em\lower.7ex\hbox{E}\kern-.125emX}}

\usepackage{float}
\usepackage{algorithm}
\usepackage{comment}
\usepackage{footnote}
\usepackage[bottom]{footmisc}
\usepackage{threeparttable}
\usepackage{multirow}
\usepackage{tabularx}
\usepackage{ragged2e}
\usepackage{booktabs}
\usepackage{makecell}
\usepackage{hyperref}
\usepackage{xspace}

\newcolumntype{C}[1]{>{\centering\arraybackslash}p{#1}}
\newcolumntype{P}[1]{>{\centering\arraybackslash}p{#1}}
\newcolumntype{M}[1]{>{\centering\arraybackslash}m{#1}}
\newcolumntype{L}[1]{>{\raggedright\arraybackslash}p{#1}}
\newcolumntype{R}[1]{>{\raggedleft\arraybackslash}p{#1}}
\newcolumntype{J}[1]{>{\justifying\arraybackslash}p{#1}}

\newcommand{\myFedDD}{\emph{FedD3}\xspace}
\newcommand{\myECEFull}{Gamma Communication Efficiency\xspace}
\newcommand{\myECEAbbr}{GCE\xspace}

\newcommand{\mypara}[1]{\noindent\textbf{#1}}

\begin{document}

\title{Federated Learning via Decentralized Dataset Distillation in Resource-Constrained Edge Environments}

\author{
    Rui Song\textsuperscript{\rm *, \rm 1, \rm 2} ,
    Dai Liu\textsuperscript{\rm *, \rm 2},
    Dave Zhenyu Chen\textsuperscript{\rm 2},\\
    \vspace{1ex}
    Andreas Festag\textsuperscript{\rm 1, \rm 3},
    Carsten Trinitis\textsuperscript{\rm 2},
    Martin Schulz\textsuperscript{\rm 2},
    Alois Knoll\textsuperscript{\rm 2} \\
    \vspace{1ex}
    $^{1}$Fraunhofer IVI \quad $^{2}$Technical University of Munich \quad $^{3}$Technische Hochschule Ingolstadt\\
    \{rui.song, andreas.festag\}@ivi.fraunhofer.de\\
    \{rui.song, dai.liu, zhenyu.chen, carsten.trinitis\}@tum.de, \\ \{schulzm, knoll\}@in.tum.de,
    andreas.festag@thi.de\\
    \thanks{$^{*}$Equal Contribution.}
    \thanks{This work was supported by the German Federal Ministry for Digital and Transport (BMVI) in the projects ``KIVI -- KI im Verkehr Ingolstadt'' and ''5GoIng – 5G Innovation Concept Ingolstadt''.}%
}

\maketitle

\input{sections/00_abstract}
\input{sections/01_intro}

\input{sections/02_background}

\input{sections/03_method}
\input{sections/05_experiment}
\input{sections/06_discussion}
\input{sections/07_conclusion}

\bibliography{ref}
\bibliographystyle{IEEEtran}

\end{document}

%% file: sections/00_abstract.tex
\begin{abstract}
In federated learning, all networked clients contribute to the model training cooperatively.
However, with model sizes increasing, even sharing the trained partial models often leads to severe communication bottlenecks in underlying networks, especially when communicated iteratively.
In this paper, we introduce a federated learning framework \myFedDD requiring only one-shot communication by integrating dataset distillation instances. 
Instead of sharing model updates in other federated learning approaches, \myFedDD allows the connected clients to distill the local datasets independently, and then aggregates those decentralized distilled datasets (e.g. a few unrecognizable images) from networks for model training.
%
%
Our experimental results show that \myFedDD significantly outperforms other federated learning frameworks in terms of needed communication volumes, while it provides the additional benefit to be able to balance the trade-off between accuracy and communication cost, depending on usage scenario or target dataset.
%
For instance, for training an AlexNet model on CIFAR-10 with 10 clients under non-independent and identically distributed (Non-IID) setting,
\myFedDD can either increase the accuracy by over 71\% with a similar communication volume, or save 98\% of communication volume, while reaching the same accuracy, compared to other one-shot federated learning approaches. 
\end{abstract}


%% file: sections/01_intro.tex
\section{Introduction}
\label{sec:intro}

\begin{figure*}[!ht]
\includegraphics[trim=0 0 0 0,clip,width=1\linewidth]{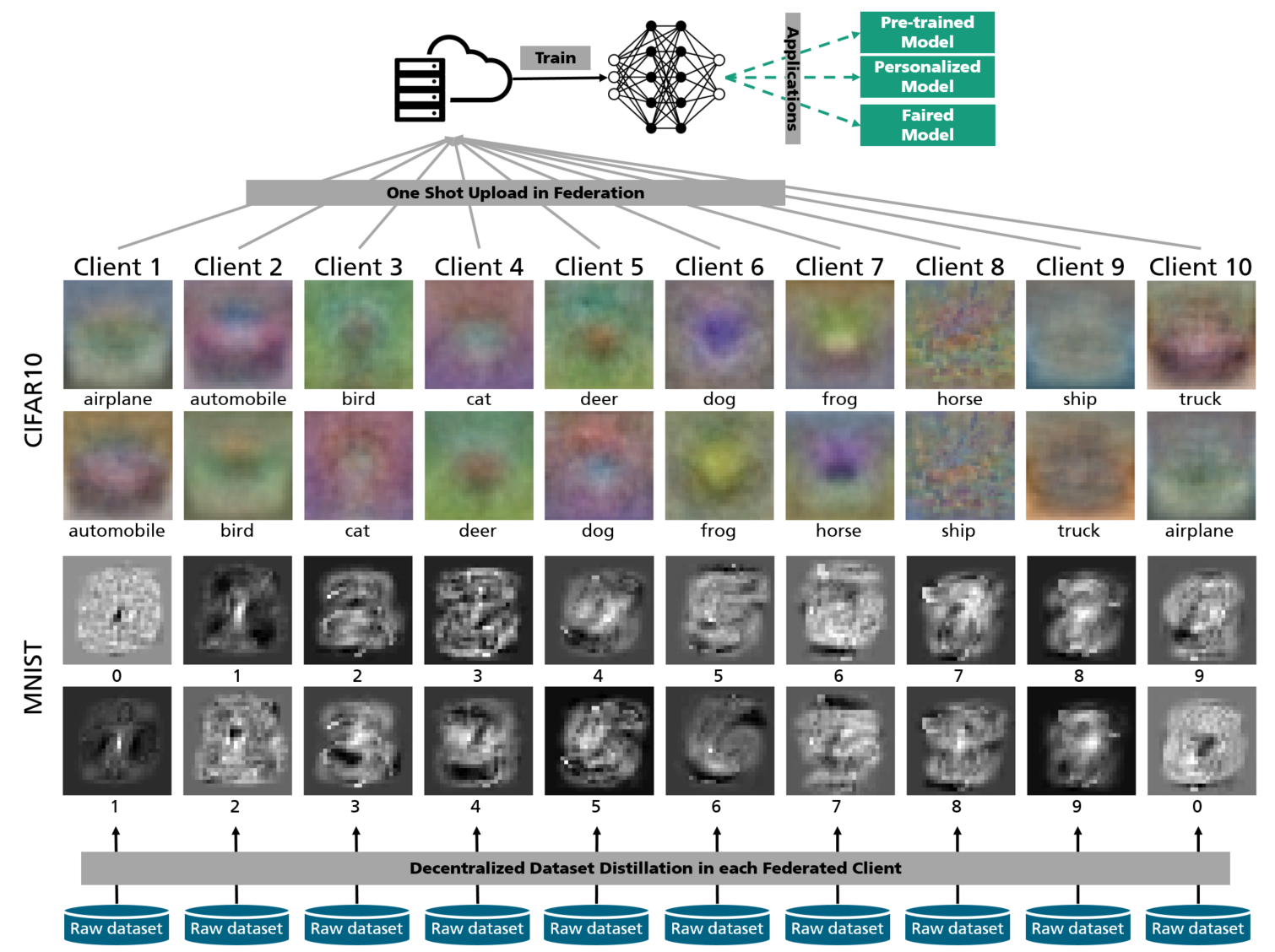}
\caption{Decentralized distilled datasets on federated CIFAR10 and MNIST divided into 10 clients. 
Each has only 2 classes of data and the class combination in each client is individual. We distill one image per class in each of 10~clients. 
}
\label{fig:cifar10_vis}
\end{figure*}

Federated learning has become an emerging paradigm for collaborative learning in large-scale distributed systems with a massive number of networked clients, such as smartphones, connected vehicles or edge devices.
Due to the limited bandwidth between clients~\cite{pmlr-v54-mcmahan17a}, previous research~\cite{pathak2020fedsplit, NEURIPS2020_39d0a890, DBLP:conf/icml/MalinovskiyKGCR20, DBLP:conf/icml/RothchildPUISB020, pmlr-v119-hamer20a, song2022resfed, gao2021convergence} attempts to speed up convergence and improve communication efficiency.
However, for modern neural networks with over hundreds of million parameters, this kind of cooperative optimization still leads to extreme communication volumes, which require substantial network bandwidth (up to the Gbps level~\cite{lin2017deep}) in order to work reliably and efficiently. 
This drawback hinders any large-scale deployment of federated learning models in commercial wireless mobile networks, e.g., vehicular communication networks~\cite{Festag:EISpringer:2015} or industrial sensor networks~\cite{9499668}.

Motivated by this communication bottleneck, prior federated learning algorithms attempt to reduce the number of communication rounds and with that the communication volume, to reach a good learning performance. 
Guha et al.~\cite{guha2019one} 
introduce a one-shot federated learning approach aimed at reducing communication overhead during the training process of a support vector machine (SVM). By exchanging information in a single communication round, it offers significant efficiency improvements.
Kasturi et al.~\cite{kasturi2020fusion} provide a fusion federated learning method that uploads both model and data distribution to the server, but characterizing a distribution of a real dataset can be difficult.
The one-shot federated learning based on knowledge transfer from Li et al.~\cite{ijcai2021-205} is general, but it requires additional communication overhead to transmit multiple student models to the server.

Inspired by the one-shot scheme~\cite{guha2019one}, we introduce a federated learning training scheme with one-shot communication via dataset distillation~\cite{wang2018dataset, cazenavette2022distillation, zhao2021dataset, nguyen2021dataset}.
Intuitively, dramatically smaller but more informative datasets, which include dense features, are synthesized and transmitted.
This way, more informative training data is transmitted across the limited bandwidth without any privacy violation.
Specifically, we introduce a novel federated learning framework incorporating dataset distillation, \myFedDD, which is shown in Fig.~\ref{fig:cifar10_vis}.
It enables efficient federated learning via transmitting the locally distilled dataset to the server in a one-shot manner, which can be applied as a pre-trained model and used for personalized~\cite{huang2021personalized, song2023fedbevt} and fairness-aware~\cite{yu2020fairness} learning.
Note that dataset distillation can keep the advantage of privacy in federated learning\cite{Dong2022privacy, sucholutsky2021secdd, li2020soft, han2020robust, wang2020federated, zhou2022communication, xiong2022feddm, li2023sharing, li2022compressed}.
It anonymously maps distilled datasets from the original client data without any exposure, which is analogous to the shared model parameters in previous federated learning methods, but substantially more efficient and effective. 

We perform an extensive analysis of our method in various scenarios to showcase the effectiveness of our method in massively distributed systems and on Non-IID (non-independent and identically distributed) datasets. 
Specifically, our experiments highlight the trade-off between accuracy and communication cost. To handle this trade-off, we propose a new evaluation metric, the $\gamma$-accuracy gain. By tuning the importance of accuracy gain $\gamma$ to the communication cost, the communication efficiency in federated learning is scored accordingly.
We also investigate the effects of specific external parameters, including Non-IID datasets, number of clients and local contributions, and demonstrate a great potential for our framework in networks with constrained communication budgets in federated learning.
We experimentally show that \myFedDD has the following advantages:
(\emph{i}) Compared to conventional multi-shot federated learning, \myFedDD~significantly reduces the amount of bits that needs to be communicated, making our approach practically feasible even in low-bandwidth environments;
(\emph{ii}) Compared to other approaches for one-shot federated learning, \myFedDD achieves a much better performance even with less communication volume, where the accuracy in a distributed system with 500~clients is enhanced by over 2.3$\times$ (from 42.08\% to 94.74\%) on Non-IID MNIST and 3.6$\times$ (from 10.74\% to 38.27\%) on Non-IID CIFAR-10 compared to \emph{FedAvg} in one single round;
(\emph{iii}) Compared to centralized dataset distillation, \myFedDD achieves much better results due to the broader data resource via federated learning.

\noindent\textbf{Contributions} 
To summarize, our contributions are as fourfold:
\begin{itemize}
\item We introduce a decentralized dataset distillation scheme in federated learning systems, where distilled data instead of models are uploaded to the server;
\item We formulate and propose a novel framework, \myFedDD, for efficient federated learning in a one-shot manner, and demonstrate \myFedDD with two different dataset distillation instances in clients;
\item We propose a novel evaluation metric $\gamma$-accuracy gain, which can be used to tune the importance of accuracy and analyze communication efficiency;
\item We conduct an extensive analysis of the proposed framework. The experiments showcase the great potentials of our framework in networks with constrained communication budget in federated learning, especially considering the trade-off between accuracy and communication costs. 
The software implementation of \myFedDD is
publicly available as open source at GitHub\footnote{\url{https://github.com/rruisong/FedD3.git}}.
\end{itemize}

%% file: sections/02_background.tex
\section{Background and Related Work}
\label{sec:background_and_related_work}

\noindent\textbf{Federated learning}
Federated learning was first introduced by McMahan et al.~\cite{pmlr-v54-mcmahan17a}, where models can be learned collaboratively from decentralized data through model exchange between clients and a central server without violating privacy. The proposed federated learning scheme \emph{FedAvg}~\cite{pmlr-v54-mcmahan17a} aggregates the received models and updates global model by averaging their parameters.

Compared to other distributed optimization approaches, federated optimization addresses more practical challenges, e.g., communication efficiency~\cite{pathak2020fedsplit}, data heterogeneity~\cite{SCAFFOLD}, privacy protection~\cite{reisizadeh2020robust}, system design~\cite{bonawitz2019towards}, which enables a large-scale deployment in real-world application scenarios~\cite{xu2023v2v4real, he2020fedml, khan2021federated, song2022itsc, chen2020scanrefer, chen2021scan2cap, chen2022d3net, xu2022opv2v, xu2021opencda, banik20213d}. 

In a federated learning scenario, given a set of clients indexed by $k$, machine learning models with weights $w_k$ are trained individually on local client datasets $\mathcal{D}_k=\{(x_i)|i=1,2,...,n_k\}$, where $x_i$ is one data point with its label $y_i$ in client $k$ and $n_k$ is the number of the local data points. 
The goal of local training in client $k$ is to minimize
\begin{equation}
    \label{Eq:Fkw}
     F_k(w) =  \frac{1}{n_k} \sum_{i=1}^{n_k} f_i(w)
\end{equation}
where $f_i(w)$ is the loss function on one data point $x_i$ with the label $y_i$.
Finally, the goal is to minimize aggregated local goals $F_k(w)$ in (\ref{Eq:Fkw}):
\begin{equation}
    \label{Eq:fw}
     f(w) = \sum_{k=1}^m \frac{n_k}{n} F_k(w)
\end{equation}
i.e. $f(w) = \mathbb{E}_{\mathcal{D}_k}[F_k(w)]$.
Note that the datasets across clients can be Non-IID in federated learning.

However, most federated optimization methods exchange models or gradients for learning updates, which can still lead to excessive communication volumes when a model has numerous parameters. This is even more problematic in wireless networks common in many mobile applications, as frequently exchanging data leads to higher error likelihood when connections are unstable, which can cause federated learning to fail.

\noindent\textbf{One-shot Federated Learning}
Federated learning with one-shot communication has been studied in several projects~\cite{guha2019one,kasturi2020fusion, ijcai2021-205}.
Guha et al.~\cite{guha2019one} introduce an algorithm for training a support vector machine in one-shot fashion.
The framework proposed by Kasturi et al.~\cite{kasturi2020fusion} uploads models and additionally the local dataset distribution, which hard to do when training on real datasets.
Li et al.~\cite{ijcai2021-205} utilize knowledge transfer to distill models in each client, which can change the original model structure and may cause much communication cost for sharing student models.

Instead of distilling models, our framework distills input data into smaller synthetic datasets in clients, which can be used for training a more general model by only relying on one-shot communication.

\noindent\textbf{Dataset Distillation}
Dataset distillation~\cite{wang2018dataset} has become an attractive paradigm.
It attempts to synthesize a significantly smaller dataset from a large dataset, aiming to maintain the same training performance in terms of test accuracy.
Dataset distillation was proposed by Wang et al.~\cite{wang2018dataset}. 
Methods like matching outputs or gradients~\cite{pmlr-v139-zhao21a, zhao2021dataset, Wang_2022_CVPR,cazenavette2022distillation} have been achieving outstanding results. 
Beside updating synthetic datasets with forward and backward propagation, Nguyen et al.~\cite{nguyen2021dataset} perform Kernel Inducing Points (KIP) and Label Solve (LS) for the optimal solution.
Relating back to our work, dataset distillation has been successfully applied in centralized training in limited fashion.

\noindent\textbf{Dataset Distillation in Federated Learning}
Though some previous works in dataset distillation, e.g.~\cite{pmlr-v139-zhao21a}, have mentioned the dataset distillation might be beneficial for federated learning, they have not provided detailed analysis and experimental evaluation on it. 
Only little research has explored the dataset distillation approaches in federated learning: Zhou et al.~\cite{zhou2020distilled} 
and Goetz et al.~\cite{goetz2020federated} have attempted to employ the approaches proposed by Wang et al.~\cite{wang2018dataset} ,and Sucholutsky and Schonlau~\cite{9533769} in federated learning, respectively. However, more advanced methods, e.g.~\cite{nguyen2021dataset}~\cite{cazenavette2022distillation} with more stable performance have been developed, and further studies on decentralized dataset distillation in federated settings are needed. Furthermore, both of them have not pointed out the dataset distillation can improve the training on heterogeneous data, which is one of the biggest challenges in federated learning.

In fact, the computation abilities in distributed edge devices are normally limited, while most dataset distillation methods require high computation power.  
For instance, the approach from~\cite{cazenavette2022distillation} can lead to computation overhead, though it can generate satisfactory distilled dataset.
Therefore, in this work, we consider the coreset-based and KIP-based~\cite{50025, nguyen2021dataset} methods for decentralized dataset distillation in our federated learning framework, and focus on improving the communication efficiency while training on federated datasets.

%% file: sections/03_method.tex
\section{FedD3: Federated Learning from Decentralized Distilled Datasets}
\label{sec:method}

To explore the dataset distillation in federated settings, we introduce \myFedDD, a federated learning framework from decentralized distilled datasets in this section. 
Specifically, we consider a joint learning task with $m$ clients, where the client $k$ owns local dataset $\mathcal{D}_k$. 
If we distill a synthetic dataset $\mathcal{ \tilde D}_{k}$ ($\tilde n_k = |\mathcal{ \tilde D}_{k}|$), in the client $k$ from its local dataset $\mathcal{D}_{k}$ ($n_k = |\mathcal{D}_{k}|$), the goal of the dataset distillation instance is to minimize $H_k(\tilde X_{k},\tilde y_{k};\Theta_k)$, where the $\tilde X_{k}$ represents the matrix of stacked distilled data points in the client $k$, $\tilde x_{k,i}$, $\tilde y_{k}$ contains the corresponding labels, and $\Theta_k$ indicates a set of parameters in the instance models. 
Note that $H_k$ can vary, depending on the instance used in the client $k$.

\input{sections/03_algo}

\subsection{Coreset-based Methods} 
We start from coreset-based methods to distill the decentralized datasets. 
We assume that there exists a synthetic dataset $\mathcal{\tilde D}_{k}$, which can approximate the statistical distribution of the original dataset $\mathcal{D}_{k}$. 
Through minimizing the error in a small subset of $\mathcal{D}_{k}$, we can generate the coreset $\mathcal{\tilde D}_{k}$ using a specific instance, e.g. Kernel Herding~\cite{10.5555/3023549.3023562}. 
More generally, we consider a clustering-based methods to generate a coreset in the client $k$ for one of classes $s\in S_k$ ($S_k$ is the set of local classes), then the goal is to minimize a clustering loss, for instance generating coreset using a Gaussian mixture model (GMM)~\cite{baudry2010combining} for each class $s$ in all clients.
%

\subsection{KIP-based Methods} 
We review and adopt KIP~\cite{50025, nguyen2021dataset} to federated fashion for its fast divide up gradient computation.
Each client $k$ aims to distill the original local dataset (a.k.a. target dataset) $\mathcal{D}$ to a synthetic dataset (a.k.a. support dataset) $\mathcal{\tilde D}$ by minimizing the kernel ridge-regression (KRR) loss $H_k(\tilde X_{k})$ as follows:
\begin{multline}
    \label{Eq:FedKIP}
     \mathrm{H_k}(\tilde X_{k}) = \\
     \frac{1}{2}||y_{k}- \mathrm{K}_k({X_{k},\tilde X_{k}}) (\mathrm{K}_k({\tilde X_{k},\tilde X_{k}}) + \lambda_k I)^{-1} \tilde y_{k}||^2_2
\end{multline}
%
where $\mathrm{K}_k$ is the kernel used in the client $k$
and $\lambda_k$ is the regularization constant deduced by ridge regression in \emph{KIP}. $I$ is an identity matrix. We refer the readers to the work of Nguyen et al.~\cite{50025} for further details regarding kernels.

\subsection{Aggregation and Learning} 
After decentralized dataset distillation, the distilled datasets in all connected clients are transmitted to the server and aggregated as $\mathcal{\tilde D} = \{\mathcal{\tilde D}_k | k=1,2,...,m\}$. 
We consider a non-convex neural network objective in the server and train a machine learning model on $\mathcal{\tilde D}$ instead of on original dataset $\mathcal{D}$, the objective is then minimizing:
\begin{equation}
    \label{Eq:FedD3}
     f(w) = \frac{1}{\tilde n}  \sum_{k=1}^m  \sum_{i=1}^{\tilde n_k} f_i (w) = \mathbb{E}_{\tilde D} [f_i(w)]
\end{equation}
where $f_i(w) = l(\tilde x_i, \tilde y_i; w)$ is the loss of the prediction on one distilled data point $\tilde x_i$ with its label $\tilde y_i$ and model weights $w$. 
If the $\tilde X_k$ could be distilled from $X_k$ perfectly, the learning result of minimizing $f(w)$ on $\tilde D$ should be similar to it on $D$. The \myFedDD pseudocode is given in Algorithm~\ref{alg:FedDD}. 
%


\subsection{\myECEFull}

In federated learning, communication cost is even more expensive than computational cost~\cite{pmlr-v54-mcmahan17a}.
It is worth studying how much communication cost is needed for achieving a dedicated gain of the model performance in terms of accuracy.
To tackle the trade-off between model performance and required communication cost, we define the \emph{\myECEFull} (\myECEAbbr) using $\gamma$-accuracy gain per binary logarithmic bit as follows:
\begin{equation}
    \label{Eq:Persuade-Accuracy Gain per byte}
      \myECEAbbr = \frac{ACC}{(1-ACC)^\gamma*\sum_{t=1}^T\log_2 (V_t+1)}
\end{equation} 
where $T\in \mathbb{N}_+$ is the total communication rounds and $V_t\in \mathbb{R}_+$ 
is the required communication volume in each round.
We use the binary logarithmic bit $\sum_{t=1}^T\log_2 (V_t+1)$ to describe the communication cost from communication round $t=1$ to $T$. 
Then the gain per binary logarithmic bit can be defined by $1/\sum_{t=1}^T\log_2 (V_t+1)$, where the 0 communication cost gives us infinitive high gain and infinitive high communication cost lead to 0 gain. 
$ACC$ is the prediction accuracy value.
$\gamma \in \mathbb{R}_+$ is an tunable parameter to represent the importance of the prediction accuracy. 
%
If $\gamma \rightarrow 0$, the accuracy and communication cost in logarithmic bit is near proportional.
The higher $\gamma$ is defined, the more test accuracy is weighted. 
A tiny test accuracy gain can be scored very well with an infinitely high $\gamma$. 
Through selecting an appropriate $\gamma$, we can evaluate the performance of federated learning approaches, considering both test accuracy and communication cost based on the application scenarios.

%% file: sections/03_algo.tex
\begin{algorithm}[t]
 \caption{\raggedright \myFedDD: Federated Learning from Decentralized Distilled Datasets}
 \label{alg:FedDD}
 \begin{algorithmic}[1]
 \renewcommand{\algorithmicrequire}{\textbf{Server Input:}}
 \renewcommand{\algorithmicensure}{\textbf{Output:}}
 \newcommand{\algorithmicbreak}{\textbf{break}}
 \newcommand{\BREAK}{\STATE \algorithmicbreak}
 \REQUIRE {number of the learning epochs $E$}
 \REQUIRE {learning rate $\eta$}
 \renewcommand{\algorithmicrequire}{\textbf{Client Input:}}
 \REQUIRE {number of the client $m$}
 \REQUIRE {number of the distillation epochs $E_k$}
 \REQUIRE {distillation rate $\eta_k$}
 \REQUIRE {decentralized dataset $\mathcal{D}_k$}
 \ENSURE  {$\tilde w$}
    \\\STATE \textbf{Server:}
    \STATE select instance with hyper-parameters $\Theta$
    \\\FOR {$k \in \{1,2,...,m\}$ \textbf{in parallel}}
        \STATE ${\tilde D_k} \leftarrow$ \textbf{ClientDatasetDistillation}(k, $\Theta$)
    \\\ENDFOR
    \\\STATE $\mathcal{\tilde D} \leftarrow aggregate(\mathcal{\tilde D}_1, \mathcal{\tilde D}_2,...,\mathcal{\tilde D}_m)$
    \\\FOR{epoch $e = 1,2,...,E$}
        \FOR{each batch $\mathcal{\tilde B} = (\tilde X_b, \tilde y_b)$ of $\mathcal{\tilde D}$}
            \STATE $\tilde w \leftarrow \tilde w - \eta (\nabla l(\tilde X_b, \tilde y_b; \tilde w))$ 
        \ENDFOR
    \ENDFOR
    \RETURN $\tilde w$
    \\\textit {~~~}
    \\\STATE \textbf{ClientDatasetDistillation}(k, $\Theta$)
    \\\STATE $\Theta_k \leftarrow \Theta$
    \\\STATE initialize $\tilde D_k$ with a subset of $D_k$ and $\tilde n_k = |\tilde D_k|$
    \\\FOR{epoch $e_k = 1,2,...,E_k$}
        \FOR{each pair of batches $\mathcal{\tilde B}_k = (\tilde X_b, \tilde y_b)$ of $\mathcal{\tilde D}_k$ and $\mathcal{B}_k = (X_b, y_b)$ of $\mathcal{D}_k$}
            \STATE ${\tilde X}_b \leftarrow {\tilde X}_b - \eta_k (\frac{\partial H_k(\tilde X_b, \tilde y_b; \Theta_k)}{\partial {\tilde X}_b})$
            \STATE update ${\mathcal{\tilde B}_k}$ and ${\mathcal{\tilde D}_k}$ with ${\tilde X}_b$
        \ENDFOR
        \IF{converged}
            \BREAK
        \ENDIF
    \ENDFOR
    \RETURN ${\tilde D_k}$
\end{algorithmic}
\end{algorithm}

%% file: sections/05_experiment.tex
\section{Experiment}
\label{sec:experiment}


\subsection{Experimental Settings}

We conduct experiments mainly on MNIST~\cite{mnist-2010} and CIFAR-10~\cite{cifar10}, as they are widely used for federated learning evaluation. 
For \myFedDD, we use GMM for coreset generation in coreset-based instance~\cite{baudry2010combining} and employ a four-layer fully connected neural network model with the width of 1024 as instance for KIP-based instance~\cite{nguyen2021dataset}.

\begin{figure*}[ht!]
\includegraphics[trim=0 0 0 0,clip,width=1\linewidth]{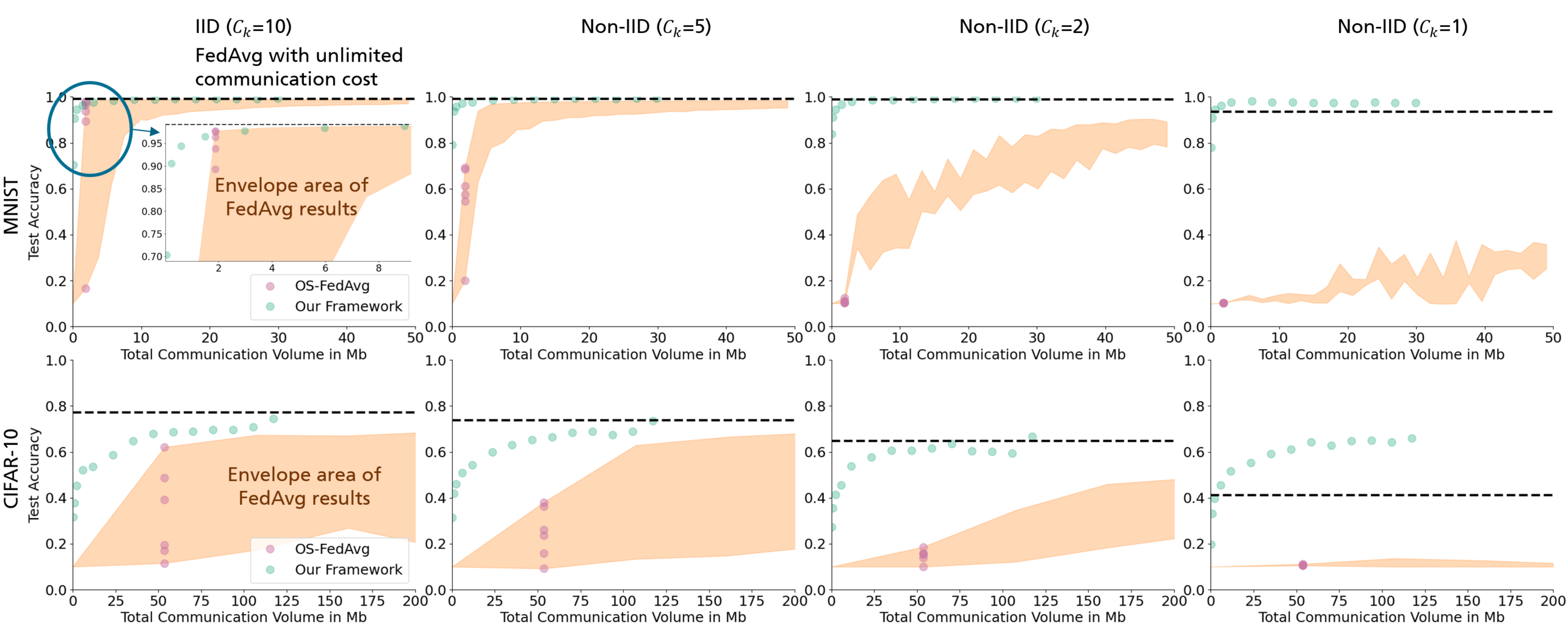}
\caption{Test accuracy changes with increasing total communication volume in Mb required in each method on MNIST and CIFAR-10. The training results are shown in two rows of four figures, where the data heterogeneity is characterized by $C_k=10, 5, 2, 1$ from left to right. We demonstrate \myFedDD with different message sizes in one-shot communication as green points. The red points are the results in one-shot FedAvg (OS-FedAvg) with different number of local epochs as baselines here. Additionally, we run standard federated learning with corresponding epochs for local training and mark the envelope area of the results in orange area. Note that the communication volume caused in \myFedDD by transmitting distilled images (1$\times$8 or 3$\times$8 bits for each gray-scale or RGB pixel), however in other federated learning by transmitting model parameters (32 bits for each parameter). For multi-shot federated learning, the total communication volume raises with increasing communication rounds.}
\label{fig:acc-msg}
\end{figure*}

\input{sections/05_table_00}
\input{sections/05_table_03}

\input{appendices/D_1_table_further_comparison}

\begin{figure}[t!]
\includegraphics[trim=0 0 0 0,clip,width=1\linewidth]{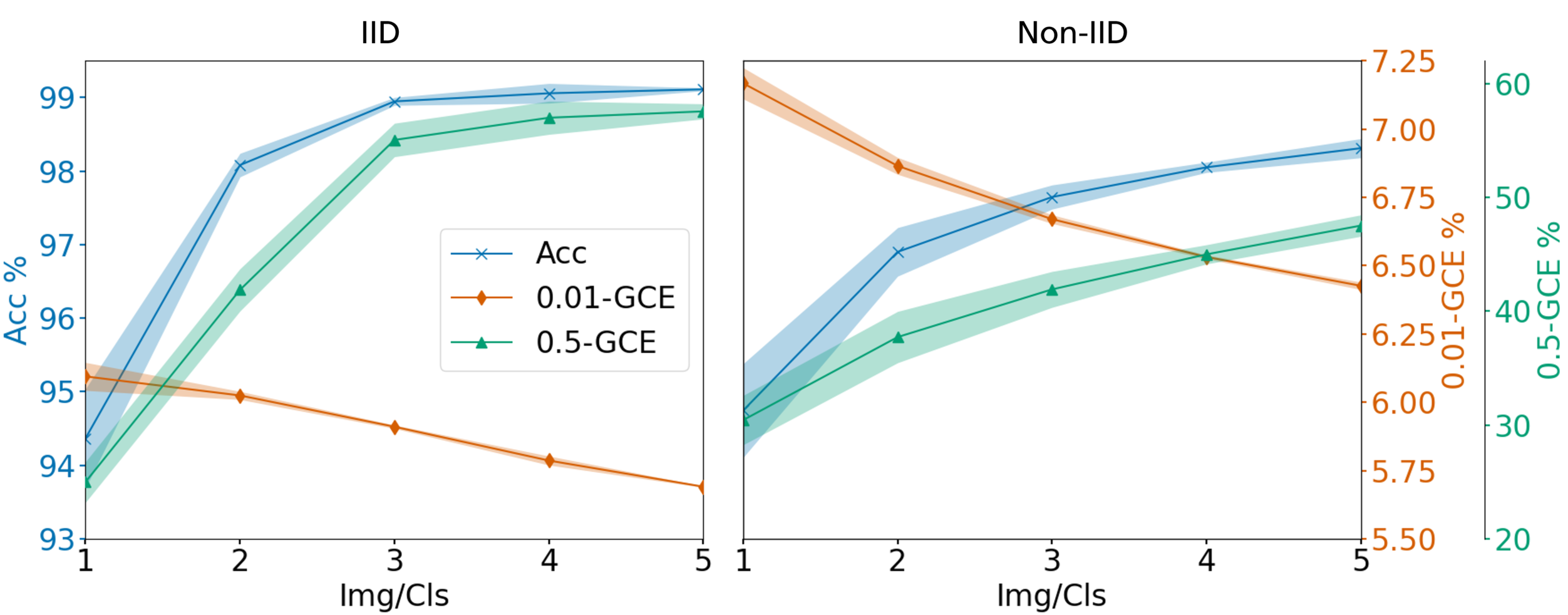}
\caption{Scalable communication efficiency by adjusting Img/Cls in decentralized dataset distillation. We set two $\gamma$ values to indicate the different importance of accuracy gain.}
\label{fig:scalable_ce}
\end{figure}

\begin{figure*}[ht!]
\centering
\includegraphics[trim=0 0 0 0,clip,width=1\linewidth]{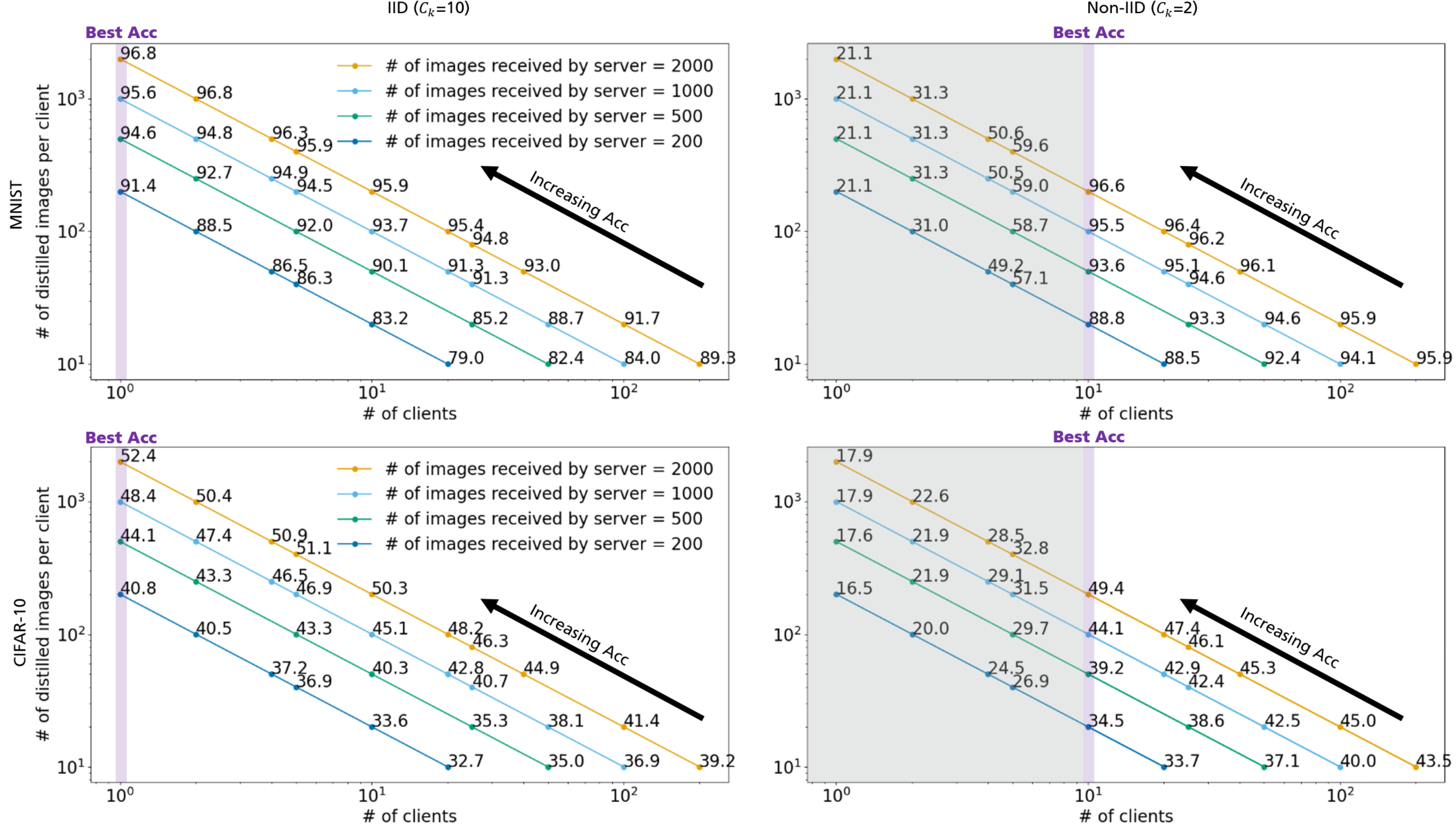}
\caption{Number of clients effects the test accuracy (in \%) of model. We consider test accuracy changing with the different number of clients. As it is naturally clear that the test accuracy of the final trained model can be improved by the increasing number of the final distilled images number at the server, through varying the number of the distilled images at each client with respect to the number of clients, we keep the size of the global distilled dataset always the same across the experiments, and hence eliminate the effects from it. Note that for Non-IID dataset, if the number of clients is less than 10 (as shown in the gray area), not all classes in the entire dataset can be covered, therefore, the test accuracy reduces. 
Apart from this impact, the results show that (\emph{i}) the test accuracy increases with smaller 
client numbers; (\emph{ii}) the test accuracy on Non-IID datasets and on IID datasets are similar, which indicates the \myFedDD is robust to data heterogeneity.
}
\label{fig:client_num_effects}
\end{figure*}

We compare \myFedDD with eight other baselines, where four federated learning methods are evaluated in both multi-shot federated learning (MSFL) and one-shot federated learning (OSFL). The federated learning methods are: FedAvg~\cite{pmlr-v54-mcmahan17a}, FedProx~\cite{li2020federated}, FedNova~\cite{wang2020tackling} and SCAFFOLD~\cite{SCAFFOLD}. The required binary logarithmic bit for communication is shown in Tab.~\ref{table:baselines}.
We train an LeNet~\cite{LeNet} and an AlexNet~\cite{alexnet} on MNIST and CIFAR-10, respectively.
Considering properties in federated learning~\cite{pmlr-v54-mcmahan17a}, we demonstrate our methods and baselines on IID and Non-IID datasets in distributed systems with a different number of clients.
We vary the number of classes in each client $C_k$ to design Non-IID datasets~\cite{li2022federated}. 
Guided by federated learning benchmark datasets made by Caldas et al. ~\cite{caldas2018leaf}, we use Non-IID to denote $C_k=2$ (pathological Non-IID~\cite{huang2021personalized}), unless specified otherwise.
%


\subsection{Robust Training on Heterogeneous Data}

First, we focus on the performance of \myFedDD on data heterogeneity and consider \emph{FedAvg} as the baseline here. 
The accuracy changing with increasing total communication volume required in each method is observed in Fig,~\ref{fig:acc-msg}. 
In experiment with IID decentralized datasets, though \emph{FedAvg} with unlimited communication cost has higher accuracy than \myFedDD,  \myFedDD outperforms \emph{FedAvg} at the same communication volume. 
The best performance of one-shot \emph{FedAvg} can be reached by using \myFedDD with only around half of the communication volume.
%
%
%
Note that we consider the communication volume only for uploading. In fact, MSFL methods still require further cost for downloading the global models.
Additionally, as shown in Fig.~\ref{fig:acc-msg}, the performance of \myFedDD is obviously more robust than \emph{FedAvg}, when the data heterogeneity across clients increases from left to right. 
With decreasing number of classes in local datasets, the prediction accuracy \emph{FedAvg} reduces notably, however, the results of \myFedDD are not much affected. 
In the scenario with extreme data heterogeneity, i.e. $C_k = 1$, the standard federated learning can even not easily converged and one-shot federated learning performances very badly, while \myFedDD can achieve the similar accuracy as in IID scenarios. 
We believe the reason is that aggregating distilled datasets allows server to train a model on a similar distribution of original datasets. 
%


\subsection{Scalable Communication Efficiency}

Then, we compare the results from 2 variants of \myFedDD with 8 baselines on both IID and Non-IID, MNIST and CIFAR10 distributed in 500 clients.
As shown in Tab.~\ref{table:comparison}, on IID datasets, \myFedDD can achieve a comparable test accuracy to other federated learning, while the \myECEAbbr is significantly higher.
On Non-IID datasets, \myFedDD is outperforms others for both test accuracy and \myECEAbbr.
We consider various $\gamma$ to indicate the evaluation based on different importance of accuracy. We assign $\gamma$ for CIFAR-10 with greater values than for MNIST, as a high accuracy for training model on CIFAR-10 is harder to achieve and should deserve to spend more communication cost for it.

We conduct further experiments 
for training a ResNet-18~\cite{he2016deep}
on Fashion-MNIST~\cite{xiao2017/online} and SVHN~\cite{netzer2011reading}, and training a CNN consisting of 5 convolutional and 3 fully connected layers
on CIFAR-100~\cite{netzer2011reading}, using MSFL and OSFL with FedAvg, and \myFedDD with KIP-instances. As shown in Tab.~\ref{table:benchmark_further}, we observe consistent results.

To perform the scalablity of communication efficiency, we meanwhile evaluate \myFedDD with increasing number of images per class in each client. 
Fig.~\ref{fig:scalable_ce} shows the test accuracy raises and thereby \myECEAbbr with higher $\gamma$ increases, when each client contributes more distilled images to the server.
However, \myECEAbbr with $\gamma=0.01$ reduces, because much communication cost is required for only a small accuracy gain.
%
%
\myFedDD provides the opportunity to optimize the \myECEAbbr by adjusting the Img/Cls in decentralized dataset distallation, considering different importance of accuracy and constraint communication budget in practical applications.


\subsection{Evaluation with System Parameters}

Finally, we explore the performance of \myFedDD affected by federated system parameters, including the number of clients and distilled images. In Fig.~\ref{fig:client_num_effects}, we run \myFedDD in federated systems containing various number of clients with IID and Non-IID decentralized datasets. 
We hold the total number of distilled images constant by varying the number of distilled images in each client. Fig.~\ref{fig:client_num_effects} shows that the prediction accuracy decreases at a larger number of clients. 
This can be led by the following reason: When increasing the client number, both the local data volume and distilled image per client reduce in our experimental setup.
This results in lower granularity and thereby decreases prediction accuracy.
In fact, federated learning considers massively distributed systems in practical applications. 
Thus, even if each client provides a small number of distilled images, a promising training performance can be achieved, when the number of clients is large.
As we can observe in Fig.~\ref{fig:client_num_effects}, when each client provides the same number of distilled images, a better model can be trained with more clients, due to more distilled images received in the server. 
Moreover, additional clients in real applications can enrich the training dataset, and hence reach a better prediction accuracy, which is consistent with the motivation of deploying federated learning.

%% file: sections/05_table_00.tex
\setlength{\tabcolsep}{3pt}
\begin{table}[t!]
\centering
\fontsize{9}{12}\selectfont
\begin{threeparttable}
\caption{The baselines and the corresponding binary logarithmic bit for one communication round, where $P$ is the bit size of models}
\label{table:baselines}

\begin{tabular}{C{1.9cm}C{1.9cm}C{1.9cm}C{1.9cm}}
\toprule 
FedAvg & FedProx & FedNova & SCAFFOLD\\

\toprule 

$\log_2 (P+1)$ & $\log_2 (P+1)$ & $\log_2 (P+9)$  & $\log_2 (2*P+1)$\\

\bottomrule 

\end{tabular}

\end{threeparttable}
\end{table}

%% file: sections/05_table_03.tex
\setlength{\tabcolsep}{3pt}
\begin{table*}[!ht]
\centering
\fontsize{7.5}{12}\selectfont
\begin{threeparttable}
\caption{The prediction accuracy (ACC) and $\gamma-$\myECEAbbr comparison between \myFedDD and other baselines on both MNIST and CIFAR-10 distributed in 500 clients. 
}
\label{table:comparison}

\begin{tabular}{C{0.2cm}C{0.3cm}|C{1.5cm}|C{1.32cm}C{1.32cm}C{1.32cm}C{1.32cm}|C{1.32cm}C{1.32cm}C{1.32cm}C{1.32cm}|C{1.32cm}C{1.32cm}}
\toprule 

\multicolumn{2}{c|}{\multirow{2}{*}{Dataset}} & \multirow{2}{*}{Metric} & \multicolumn{4}{c|}{MSFL\tnote{1}} & \multicolumn{4}{c|}{OSFL} & \multicolumn{2}{c}{FedD3 (Ours)\tnote{2}}\\

&&& {FedAvg} & {FedProx} & {FedNova} & {SCAFFOLD} & {FedAvg} & {FedProx} & {FedNova} & {SCAFFOLD} & {Coreset} & {KIP}\\

\midrule 
\midrule 

\multirow{6}{*}{\rotatebox[origin=c]{90}{MNIST}} & \multirow{3}{*}{\rotatebox[origin=c]{90}{IID}} & {ACC \%} & {96.97$\pm$0.02} & {96.55$\pm$0.03} & {85.50$\pm$0.01} & \textbf{97.49$\pm$0.02} & {85.34$\pm$0.02} & {83.63$\pm$0.02} & {74.78$\pm$0.02} & {67.47$\pm$5.89} & {86.82$\pm$0.46} & \textbf{94.37$\pm$0.67}\\

&& {0.01-\myECEAbbr \%} & {0.27$\pm$0.01} & {0.27$\pm$0.00} & {0.69$\pm$0.00} & {0.13$\pm$0.00} & {4.16$\pm$0.00} & {4.07$\pm$0.00} & {3.63$\pm$0.00} & {1.63$\pm$0.15} & {5.56$\pm$0.03} & \textbf{6.09$\pm$0.05}\\

&& {0.5-\myECEAbbr \%} & {1.48$\pm$0.00} & {1.38$\pm$0.01} & {1.79$\pm$0.00} & {0.82$\pm$0.00} & {10.66$\pm$0.01} & {9.88$\pm$0.01} & {7.12$\pm$0.01} & {2.85$\pm$0.54}  & {15.01$\pm$0.34} & \textbf{24.99$\pm$1.78 }\\

 \cmidrule(lr{0em}){2-13}

& \multirow{3}{*}{\rotatebox[origin=c]{90}{Non-IID}} & {ACC \%} & {71.29$\pm$0.02} & {67.54$\pm$0.04} & {71.33$\pm$0.08} & {88.69$\pm$0.12} & {42.08$\pm$0.03} & {49.25$\pm$0.01} & {63.61$\pm$0.75} & {36.92$\pm$0.03} & {77.29$\pm$2.58} & \textbf{94.74$\pm$0.64}\\

&& {0.01-\myECEAbbr \%} & {0.20$\pm$0.00} & {0.20$\pm$0.00} & {0.20$\pm$0.00} & {0.12$\pm$0.00} & {2.02$\pm$0.00} & {2.37$\pm$0.00} & {3.07$\pm$0.04} & {0.89$\pm$0.00} & {5.76$\pm$0.20} & \textbf{7.17$\pm$0.06}\\

&& {0.5-\myECEAbbr \%} & {0.31$\pm$0.00} & {0.31$\pm$0.00} & {0.30$\pm$0.00} & {0.35$\pm$0.00} & {2.64$\pm$0.00} & {3.31$\pm$0.00} & {5.04$\pm$0.11} & {1.11$\pm$0.00} & {11.95$\pm$1.13} & \textbf{30.43$\pm$2.17}\\

\midrule 
\midrule 

\multirow{6}{*}{\rotatebox[origin=c]{90}{CIFAR-10}} & \multirow{3}{*}{\rotatebox[origin=c]{90}{IID}} & {ACC \%} & \textbf{48.12$\pm$0.38} & {47.89$\pm$0.27} & {47.69$\pm$0.80} & {41.89$\pm$0.11 } & {36.23$\pm$0.60} & {36.26$\pm$0.63} & {36.59$\pm$0.27} & {24.33$\pm$0.01} & {46.18$\pm$0.08} & \textbf{48.97$\pm$0.83}\\

&& {0.1-\myECEAbbr \%} & {0.60$\pm$0.76} & {0.33$\pm$0.00} & {0.42$\pm$0.06} & {0.14$\pm$0.00} & {1.47$\pm$0.03} & {1.47$\pm$0.03} & {1.49$\pm$0.01} & {0.49$\pm$0.00} & {2.74$\pm$0.01} & \textbf{2.92$\pm$0.05}\\

&& {2-\myECEAbbr \%} & {1.60$\pm$1.35} & {1.14$\pm$0.02} & {1.34$\pm$0.21} & {0.40$\pm$0.00} & {3.46$\pm$0.12} & {3.47$\pm$0.13} & {3.54$\pm$0.06} & {0.83$\pm$0.00} & {8.91$\pm$0.04} & \textbf{10.50$\pm$0.53}\\

 \cmidrule(lr{0em}){2-13}

& \multirow{3}{*}{\rotatebox[origin=c]{90}{Non-IID}} & {ACC \%} & {13.14$\pm$5.02} & {18.46$\pm$0.83} & {12.98$\pm$5.00} & {34.27$\pm$0.04} & {10.73$\pm$0.01} & {10.71$\pm$0.01} & {10.72$\pm$0.01 } & {10.05$\pm$0.00} & {30.32$\pm$1.43} & \textbf{38.27$\pm$1.45}\\

&& {0.1-\myECEAbbr \%} & {0.35$\pm$0.04 } & {0.19$\pm$0.04} & {0.39$\pm$0.00} & {0.12$\pm$0.00} & {0.42$\pm$0.00} & {0.42$\pm$0.00} & {0.42$\pm$0.00} & {0.20$\pm$0.00} & {2.02$\pm$0.10} & \textbf{2.58$\pm$0.10}\\

&& {2-\myECEAbbr \%} & {0.44$\pm$0.04} & {0.24$\pm$0.05} & {0.48$\pm$0.00} & {0.26$\pm$0.00} & {0.52$\pm$0.00} & {0.52$\pm$0.00} & {0.52$\pm$0.00} & {0.24$\pm$0.00} & {4.01$\pm$0.36} & \textbf{6.45$\pm$0.56}\\

\bottomrule 
\end{tabular}
\begin{tablenotes}
    \scriptsize 
     \item[1] We select the best result in the first 18 and 6 communication rounds for the training on MNIST and CIFAR-10, respectively.
     \item[2] Each client contributes 1 distilled image per class (Img/Cls = 1) from its local dataset. 
   \end{tablenotes}
\end{threeparttable}
\end{table*}

%% file: appendices/D_1_table_further_comparison.tex
\setlength{\tabcolsep}{3pt}
\begin{table}[ht!]
\centering
\fontsize{8}{12}\selectfont
\begin{threeparttable}
\caption{The prediction accuracy (Acc) and $\gamma-$\myECEAbbr comparison between \myFedDD and other baselines on Fashion-MNIST and CIFAR-100 distributed in 200 clients, and on SVHN distributed in 100 clients.}
\label{table:benchmark_further}

\begin{tabular}{C{0.2cm}C{0.3cm}|C{1.7cm}|C{1.42cm}C{1.42cm}C{1.42cm}}
\toprule 

\multicolumn{2}{c|}{\multirow{1}{*}{Dataset}} & {Metric} & \multicolumn{1}{c}{MSFL} & \multicolumn{1}{c}{OSFL} & \multicolumn{1}{c}{FedD3}\\

\midrule 
\midrule 

\multirow{6}{*}{\rotatebox[origin=c]{90}{Fashion-MNIST}} & \multirow{3}{*}{\rotatebox[origin=c]{90}{IID}} & {Acc \%} & \textbf{79.40$\pm$0.00} & {69.69$\pm$0.07} & \textbf{74.80$\pm$0.75}\\

&& {0.01-\myECEAbbr\%} & {0.95$\pm$0.00} & {2.48$\pm$0.00} & \textbf{4.65$\pm$0.05}\\

&& {0.5-\myECEAbbr\%} & {2.05$\pm$0.00} & {4.45$\pm$0.01} & \textbf{9.13$\pm$0.23}\\

 \cmidrule(lr{0em}){2-6}

& \multirow{3}{*}{\rotatebox[origin=c]{90}{Non-IID}} & {Acc \%} & {40.69$\pm$0.01} & {27.92$\pm$0.00} & \textbf{76.78$\pm$0.98}\\

&& {0.01-\myECEAbbr\%} & {0.48$\pm$0.00} & {0.99$\pm$0.00} & \textbf{5.56$\pm$0.07}\\

&& {0.5-\myECEAbbr\%} & {0.62$\pm$0.00} & {1.16$\pm$0.00} & \textbf{11.39$\pm$0.39}\\

\midrule 
\midrule 

\multirow{6}{*}{\rotatebox[origin=c]{90}{SVHN}} & \multirow{3}{*}{\rotatebox[origin=c]{90}{IID}} & {Acc \%} & \textbf{80.99$\pm$0.00} & {25.00$\pm$0.49} & \textbf{80.42$\pm$0.63}\\

&& {0.01-\myECEAbbr\%} & {0.16$\pm$0.00} & {0.44$\pm$0.01} & \textbf{3.85$\pm$0.03}\\

&& {0.5-\myECEAbbr\%} & {0.36$\pm$0.00} & {0.51$\pm$0.01} & \textbf{8.56$\pm$0.21}\\

 \cmidrule(lr{0em}){2-6}

& \multirow{3}{*}{\rotatebox[origin=c]{90}{Non-IID}} & {Acc \%} & {46.32$\pm$0.06} & {19.96$\pm$0.05} & \textbf{69.10$\pm$0.98}\\

&& {0.01-\myECEAbbr\%} & {0.10$\pm$0.01} & {0.35$\pm$0.00} & \textbf{3.69$\pm$0.05}\\

&& {0.5-\myECEAbbr\%} & {0.13$\pm$0.02} & {0.39$\pm$0.01} & \textbf{6.56$\pm$0.02}\\

\midrule 
\midrule 

\multirow{6}{*}{\rotatebox[origin=c]{90}{CIFAR-100}} & \multirow{3}{*}{\rotatebox[origin=c]{90}{IID}} & {Acc \%} & \textbf{41.15$\pm$0.03} & {20.21$\pm$0.06} & \textbf{47.89$\pm$0.42}\\

&& {0.1-\myECEAbbr\%} & {0.15$\pm$0.01} & {0.36$\pm$0.00} & \textbf{2.41$\pm$0.02}\\

&& {2-\myECEAbbr\%} & {0.05$\pm$0.00} & {0.55$\pm$0.00} & \textbf{8.31$\pm$0.21}\\

 \cmidrule(lr{0em}){2-6}

& \multirow{3}{*}{\rotatebox[origin=c]{90}{Non-IID\tnote{1}}} & {Acc \%} & {29.79$\pm$0.03} & {10.27$\pm$0.00} & \textbf{38.05$\pm$0.49}\\

&& {0.1-\myECEAbbr\%} & {0.04$\pm$0.00} & {0.18$\pm$0.00} & \textbf{1.97$\pm$0.03}\\

&& {2-\myECEAbbr\%} & {0.07$\pm$0.00} & {0.22$\pm$0.00} & \textbf{4.90$\pm$0.14}\\

\bottomrule 

\end{tabular}
\begin{tablenotes}
     \scriptsize
     \item[1] Each client holds data with 50 classes ($C_k=50$).
   \end{tablenotes}
\end{threeparttable}
\end{table}

%% file: sections/06_discussion.tex
\section{Discussion}
\label{sec:discussion}


\mypara{Distilled Datasets in Multi-round}
Due to the robustness on data heterogeneity, we further explore the potential benefits of distilled datasets in federated learning. 
We believe that sharing such synthetic data might bridge the information silos. 
For that, we extend \myFedDD to multiple shots and consider a hybrid federated learning method by adding a spoon of distilled datasets from other clients via D2N (Device to Networks) or D2D (Device to Device) networks, before the first round in standard federated learning. 

\begin{figure}[t!]
\centering
\includegraphics[trim=0 0 0 0,clip,width=0.9\linewidth]{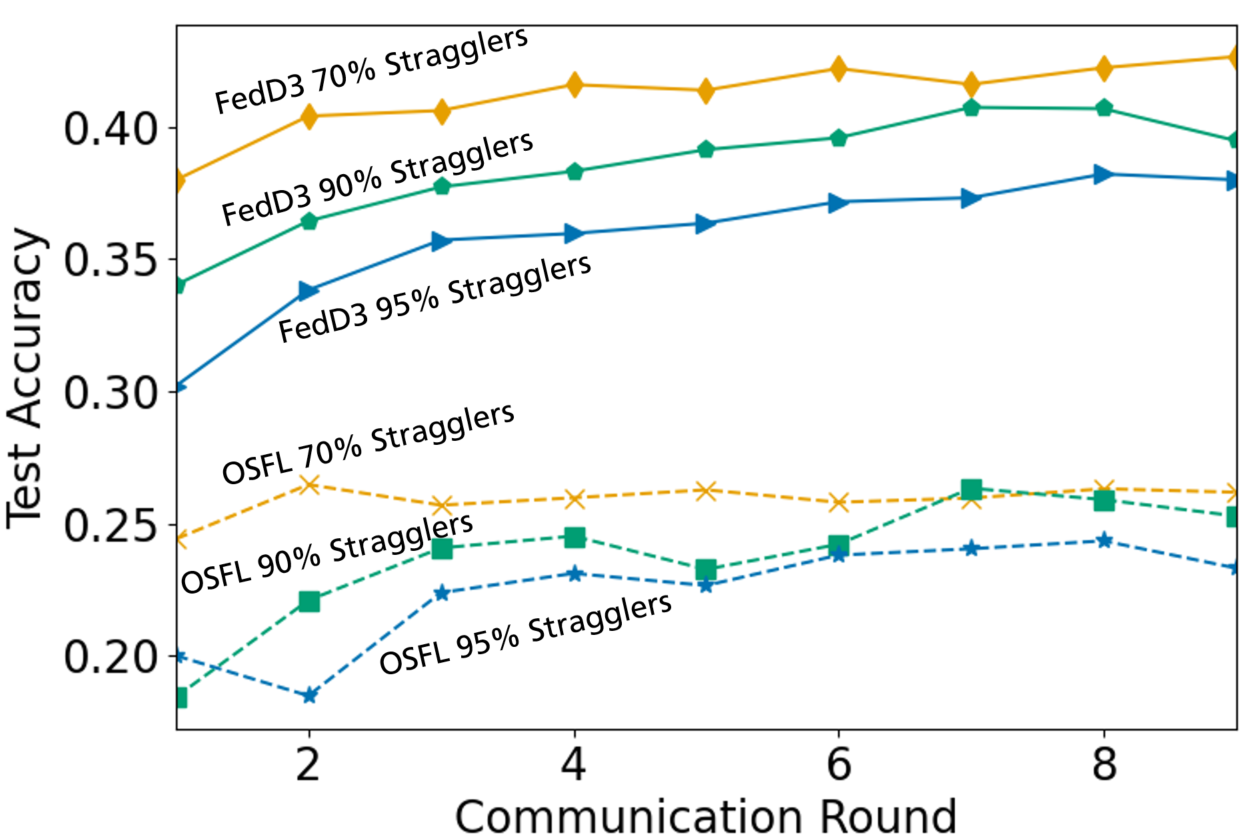}
\caption{Performance comparison of \myFedDD and OSFL for training a CNN model on distributed Non-IID CIFAR-100 in heterogeneous networks (with different ratios of stragglers).}
\label{fig:network}
\end{figure}

\mypara{Network Assumption} 
Compared to multi-shot schemes, a one-shot scheme is less affected by network heterogeneity. Despite, to evaluate the impact of network heterogeneity and address the network assumption in federated learning application scenarios, we consider the Quality of Service (QoS) of \emph{at most once} for the one-shot scheme and compare the performance of \myFedDD and OSFL with different ratios of stragglers. As shown in Fig~\ref{fig:network}, FedD3 outperforms OSFL.

\mypara{Computation Costs}
Most dataset distillation methods can result in high computation loads. In our experiment setup, we take into account computation limitations in KIP and scale the number of distilled images on each client. In our \myFedDD setup, we use FC-4 kernel without pooling and convolutional kernels for dataset distillation. In each client, we only distill a small number of images, i.e. 1 Img/Cls. In Federated Learning, local training can result in a linear increase in computational cost with the number of epochs. We also disregard the computational cost caused by a large number of epochs and aim to achieve better results in the baselines for a fair comparison only for communication efficiency. Additionally, we measure the required time on a compact NVIDIA Quadro RTX 4000 GPU for training an AlexNet on CIFAR-10: On average \emph{73.69s} are needed for \myFedDD with 800 distillation steps for 1 Img/Cls, and \emph{89.96s} for one-shot FedAvg with 10 epochs.

\mypara{Beyond Kernel: Individual DD-instances}
Because of stable performance, we mainly employ KIP methods as dataset distillation instances in \myFedDD. 
In fact, according to the actual local datasets, we can also consider individual instances to generate the synthetic dataset for uploading. 
Self-selecting instances in each client should be encouraged, which can be advantageous to the quality of distilled data, as the server and other clients should be not aware of the distribution of local data. 
Additionally, we believe the autonomy of local dataset distillation can enhance privacy.


\mypara{Lessons Learned for Orchestration}
In \myFedDD, all local datasets should meet the requirement of the used dataset distillation method. Here we can think about two example scenarios: 
(\emph{i}) If there is only one data point in one of the clients, $\tilde D_k$ and $D_k$ are the same at the first epoch in \emph{ClientDatasetDistillation} in Algorithm~\ref{alg:FedDD};
(\emph{ii}) If there is only one data point with its label $x_i$ for a specific class $y_c$, the distilled dataset $\tilde x_i$ can be always the same as $x_i$, especially when only the loss is backpropagated on $\tilde X$. 
Both situations can break the privacy, as it will upload at least one raw data point to the server. 
To tackle this issue, we believe a good orchestration for dataset and client selection is recommended. 

%% file: sections/07_conclusion.tex
\section{Conclusion}
\label{sec:conclusion}

In this work, we introduce a novel federated learning framework, \myFedDD, which reduces the overall communication volume and with that opens up the concept of federated learning to more application scenarios in particular in network constrained environments.
It achieves this by leveraging local dataset distillation instead of traditional learning approaches \emph{(i)} to significantly reduce communication volumes and \emph{(ii)} to limit transfers to one-shot communication, rather than iterative multi-way communication.
%
%
Our conducted experimental evaluation shows that \myFedDD can well balance the trade-off between prediction accuracy and communication cost for federated learning. 
Compared to other federated learning, it provides a more robust training performance, especially on Non-IID data silos. 
